\documentclass[11pt]{article}
\usepackage{acl}
\usepackage{graphicx}
\usepackage{booktabs}
\usepackage{amsmath}
\usepackage{multirow}
\usepackage{xcolor}

\title{Scaling Enterprise Agent Routing: \\
Degradation, Diagnosis, and Recovery}

\author{Kellen Gillespie \and Robyn Perry \\
  Superhuman, Inc. \\
  \texttt{\{kellen.gillespie, robyn.perry\}@grammarly.com}}

\begin{document}
\raggedbottom
\maketitle

\begin{abstract}

Production LLM assistants route user requests to growing libraries of specialized tools, but how does routing accuracy degrade as the catalog scales?
We study single-step routing on a 110-agent, 584-tool catalog from a deployed enterprise productivity assistant, evaluating three frontier models from 10 to 110 agents.
Routing F1 on under-specified requests drops 16--23 percentage points across models.
An oracle analysis decomposes the degradation into a \emph{retrieval} gap (the model cannot surface the right tool) and a \emph{confusion} gap (even with perfect retrieval, the oracle ceiling drops 10pp).
Embedding-based shortlisting recovers +10--11pp F1 at full scale across all three models and two providers.
A production annotation study (1,435 human-labeled utterances, three annotators) confirms the recovery on real traffic at +10--17pp despite 10--15pp lower absolute performance.

\end{abstract}

\section{Introduction}
\label{sec:intro}

LLM-based assistants increasingly serve as orchestration layers that route user requests to specialized agents for email, project tracking, scheduling, and more.
As organizations add agents to these systems, the routing decision becomes harder and the model must select from a growing catalog of semantically overlapping options.

This scaling challenge is already driving platform-level responses. OpenAI introduced namespace-based tool search, Anthropic provides BM25 retrieval over tool descriptions, and MCP server registries are growing beyond what flat tool lists can support.
Prior work shows that tool-calling performance degrades with catalog size \cite{longfunceval} and that retrieval errors dominate agent failures \cite{livemcpbench}, but the mechanism (what breaks, at what scale, and what levers exist) remains undercharacterized.

We present a controlled study of single-step routing accuracy across 10--110 agents on a production-sourced catalog from a deployed enterprise productivity assistant.
Our analysis has two parts:

\begin{enumerate}
    \item \textbf{Scaling diagnosis} (\S\ref{sec:scaling-curves}). F1 drops 16--23pp, driven primarily by recall. An oracle analysis decomposes this into a \emph{retrieval} gap (the model cannot surface the right tool) and a \emph{confusion} gap (the oracle ceiling drops from 79\% to 69\%). The confusion gap is amplified by the enterprise productivity domain, where functionally similar tools (Gmail/Outlook for email, Improve/Paraphraser/Proofreader for writing, Jira/Asana for project management) grow naturally with the catalog \cite{toolbench,toolret,bfcl}.

    \item \textbf{Shortlisting as intervention} (\S\ref{sec:shortlisting}, \S\ref{sec:production-validation}). Embedding shortlisting recovers +10--11pp F1 at full scale across three models from two providers. The recovery holds on 1,435 human-annotated production utterances (+10--17pp). Tool-level retrieval outperforms all pack-level approaches (hierarchical LLM routing, pack-level embedding, platform tool search) by 2--4pp. An error composition analysis (\S\ref{sec:error-analysis}) shows that shortlisting cuts routing misses from 31\% to 10\% at the cost of a stable 9\% shortlister miss rate.
\end{enumerate}

\section{Related Work}
\label{sec:related}

\paragraph{Tool-count scaling.}
\citet{longfunceval} stress-test tool calling at 49--741 tools, reporting 7--85\% performance drops.
LiveMCPBench \cite{livemcpbench} finds retrieval errors account for ${\sim}$50\% of agent failures across 527 tools.
Toolshed \cite{toolshed}, ScaleMCP \cite{scalemcp}, MonoScale \cite{monoscale}, and RAG-MCP \cite{ragmcp} document performance collapse with growing tool and agent pools.
We add a precision/recall decomposition and controlled mitigations to this line of work.

\paragraph{Tool retrieval and selection.}
Toolformer \cite{toolformer} teaches LMs to insert tool calls during generation.
Retrieve-then-route approaches range from document retrieval over API catalogs \cite{gorilla,toolbench}, fine-tuned retrievers \cite{toolret,skillrouter}, reranking and query rewriting \cite{toolrerank,reinvoke}, and token-level tool encoding \cite{toolkengpt,toolgen}.
Benchmarks and data generation pipelines \cite{apigen,sealtools,toolde} complement these with evaluation methodology.
ToolScope \cite{toolscope} tackles semantic overlap by merging similar tools at the catalog level.
We show that dense embedding retrieval outperforms both platform approaches and a fine-tuned retriever \cite{toolret} without an LLM call.

\paragraph{Agent system scaling.}
\citet{kim2026scaling} study when multi-agent coordination outperforms single agents, while AgentArch \cite{agentarch} varies agent architecture on fixed tool sets.
HuggingGPT \cite{hugginggpt} and AnyTool \cite{anytool} route via hierarchical dispatch over models and API tiers respectively.
ScaleCall \cite{scalecall} evaluates hybrid retrieval for enterprise tool selection.
We hold architecture constant and vary catalog size, comparing hierarchical dispatch against flat embedding retrieval.

\section{Experimental Setup}
\label{sec:setup}

\subsection{Agent Catalog}

Our catalog comprises 110 agents and 584 individual tools from a deployed enterprise productivity assistant, ranging from single-purpose agents (Weather, Google Translate) to multi-tool suites (Gmail with 15+ actions, Jira with 20+).
The catalog has natural semantic overlap (multiple email clients, writing tools, project trackers, and document editors), creating routing ambiguity uncommon in API-centric benchmarks \cite{toolbench,toolret,bfcl}.
We evaluate \textbf{minimal} (name and description) and \textbf{rich} (name, description, examples, semantic tags, enriched descriptions) metadata variants.

\subsection{Evaluation Data}

\paragraph{Synthetic queries.}
4,105 queries generated by GPT-4o across difficulty levels: \emph{explicit} (names the tool: ``send a Gmail'') and \emph{implicit} (describes the need without naming it: ``email the team about Monday's deadline'').
Each query has a target tool and \emph{also-valid} labels enabling dynamic ground truth that adjusts to each sampled tool set.
The routing models (GPT-5.x, Sonnet) differ from the generation model, though GPT-4o and GPT-5.x share a provider, creating potential distributional affinity. Cross-provider replication with Sonnet and the production validation (\S\ref{sec:production-validation}) on human-written utterances mitigate this concern.

\paragraph{Production queries.}
1,435 utterances sampled from production traffic of the deployed system, stratified across agents with sufficient traffic (capped at 100 utterances per agent) and filtered for quality (language, safety).
Three trained linguists independently annotated each utterance with top-5 candidates from an LLM-based shortlister (GPT-5.4), independent of the embedding retriever evaluated in \S\ref{sec:shortlisting}. Each annotator rated every candidate as \emph{best option}, \emph{also valid}, or \emph{not applicable}, and could nominate tools outside the shortlisted set; fewer than 1\% of gold labels (13 cases) required out-of-pool nominations. Gold labels use majority vote ($\geq$2/3 annotators).
Per-candidate Krippendorff's $\alpha{=}0.68$ (ordinal), reflecting the catalog's semantic overlap: annotators agree on which tools are relevant but often disagree on which is \emph{best} among near-equivalents. At the item level, 94\% of utterances have at least one shared valid tool across annotators.
All production queries are implicit.

\paragraph{Implicit queries as primary metric.}
Explicit queries are near-ceiling ($>$90\% F1) at all scales.
We report on implicit queries throughout, as they represent realistic production traffic where the user does not name the target tool.

\subsection{Models and Routing}

We evaluate three frontier models from two providers: \textbf{GPT-5.1} and \textbf{GPT-5.4} (OpenAI, function calling via Responses API) and \textbf{Claude Sonnet 4.5} (Anthropic, native tool use).
All use function-calling interfaces where the tool catalog is provided as callable function definitions.

\subsection{Scale Points and Sampling}

We evaluate at scale points of 10, 20, 30, 40, 60, 80, 100, and 110 agents (51--584 tools).
At each non-endpoint scale, we sample $k{=}3$ agent subsets (folds) and report fold-averaged metrics with bootstrap 95\% confidence intervals.
The 110-agent endpoint is the full catalog (single subset), so CIs there are query-level only.
Infrastructure agents (general assistant, knowledge search, web search) are always present.

\subsection{Metrics}

Multi-label precision, recall, and F1, computed per-query against the dynamic valid set (target tool plus any also-valid tools present in the current fold).

\section{How Routing Degrades at Scale}
\label{sec:scaling}

\subsection{Scaling Curves}
\label{sec:scaling-curves}

\begin{figure*}[t]
\centering
\includegraphics[width=0.8\textwidth]{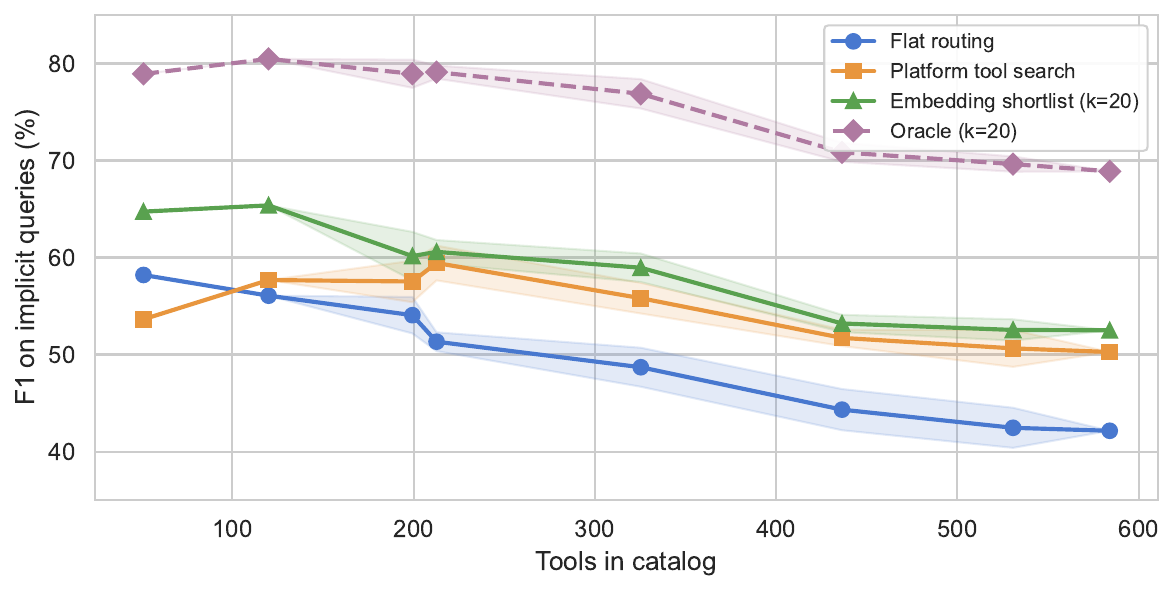}
\caption{Routing F1 (implicit queries) across catalog scale (GPT-5.4). Shaded bands: fold standard deviation. Flat routing degrades from 58\% to 42\%. Embedding shortlisting ($k{=}20$) recovers +10pp at full scale. Oracle ceiling (dashed) drops 10pp, indicating confusion independent of retrieval quality. Tool search helps above $\sim$180 tools but plateaus.}
\label{fig:unified-scaling}
\end{figure*}

Figure~\ref{fig:unified-scaling} shows routing F1 on implicit queries as catalog size grows from 51 to 584 tools.
Flat tool-level routing (GPT-5.4) drops from 58.2\% to 42.1\%.
The degradation is \emph{recall-driven}: precision drops moderately (68\%$\rightarrow$60\%) while recall drops more than twice as fast (55\%$\rightarrow$37\%).
As the catalog grows, the model increasingly misses valid tools more often than it selects invalid ones.
A fixed cohort of 731 queries present at every scale point shows the same degradation (14.9pp), confirming it is driven by catalog growth rather than query composition (Appendix~\ref{app:fixed-cohort}).

\paragraph{Two-component decomposition.}
An oracle shortlister (all dynamically-valid tools plus random distractors to fill 20 slots) establishes an upper bound on what routing can achieve with perfect retrieval.
The oracle drops from 79.0\% to 68.8\%, a 10pp decline even with the correct tool always present.
This reveals two independent degradation sources:
(a)~a \emph{retrieval} gap, the 16pp difference between oracle and practical shortlisting at full scale, recoverable by better retrievers; and
(b)~a \emph{confusion} gap, the oracle's own 10pp decline.
This decline reflects both incomplete coverage of growing equivalence classes (valid-set size increases from 1.6 to 3.2 tools per query at scale) and genuine inter-tool confusion. Valid-set growth affects recall but not precision, so the 8pp precision drop (68\%$\rightarrow$60\%) confirms confusion independent of the coverage effect.
The effective confusion gap in practice is likely larger than 10pp, since the oracle's random distractors are less confusable than a real retriever's semantically similar candidates.

\paragraph{Cross-model reproducibility.}
\label{sec:model-universality}
All three models exhibit the same recall-driven degradation pattern with an elbow at 40--60 agents (Appendix~\ref{app:full-results}).
GPT-5.4 outperforms GPT-5.1 by 4--8pp across scale. Sonnet starts higher (66.3\% at 51 tools) but degrades faster ($-$20pp vs.\ $-$16pp for GPT-5.4).
Stronger models provide a constant offset but follow the same curve.

\paragraph{Metadata and tool search.}
\label{sec:granularity-metadata}
Rich metadata (examples, tags, enriched descriptions) provides a scale-increasing benefit at the agent level ($+$1.2pp at 20, $+$4.2pp at 110) but near-zero effect at the tool level ($<$1pp).
Metadata quality complements architectural changes but does not substitute for them.
OpenAI's namespace-based tool search provides partial relief at moderate scale but plateaus at larger catalogs (\S\ref{sec:retrieval-comparison}).

\section{Shortlisting Across Scale}
\label{sec:shortlisting}

We ask whether pre-filtering the catalog to a small shortlist can close the retrieval gap responsible for most of the scaling loss.

\subsection{Shortlisting Comparison}
\label{sec:unified-comparison}

Figure~\ref{fig:unified-scaling} compares four approaches across 51--584 tools.
Embedding shortlisting (text-embedding-3-large, $k{=}20$ tools) outperforms flat routing at every scale point ($+$6--11pp, paired bootstrap $p{<}0.01$ at all points) and matches or exceeds platform tool search throughout.
We fix $k{=}20$ based on a sensitivity analysis (Appendix~\ref{app:k-sensitivity}) showing that F1 plateaus at $k{\geq}10$ and is statistically indistinguishable from $k{=}20$ to $k{=}50$.
At 584 tools embedding reaches 52.5\%, while tool search reaches 50.3\% and flat (no shortlisting) reaches 42.1\%.

The 16pp gap between embedding and oracle at full scale reflects retrieval quality, since embedding returns semantically similar distractors that are harder for the router than random ones.
Across approaches, even oracle recall drops 15pp (Figure~\ref{fig:precision-recall}, Appendix~\ref{app:full-results}).

\subsection{Cross-Model Reproducibility}
\label{sec:cross-model}

Table~\ref{tab:cross-model} shows the shortlisting recovery across models and providers.
All three converge to ${\sim}$+10pp at full scale despite different baselines (GPT-5.4: 42.1\%, GPT-5.1: 40.6\%, Sonnet: 45.9\%).
Sonnet's smaller delta at 120 tools ($+$2.7pp) reflects its stronger baseline, leaving less recall to recover.
As Sonnet's baseline degrades at larger catalogs, the shortlisting benefit grows, matching the GPT models at full scale.

\begin{table}[t]
\centering
\small
\begin{tabular}{lccc}
\toprule
& \textbf{120} & \textbf{325} & \textbf{584} \\
\textbf{Model} & \textbf{tools} & \textbf{tools} & \textbf{tools} \\
\midrule
GPT-5.4 flat & 56.1 & 48.7 & 42.1 \\
\quad + emb $k{=}20$ & 65.4 & 58.9 & 52.5 \\
\quad $\Delta$ & +9.3 & +10.3 & +10.4 \\
\midrule
GPT-5.1 flat & 57.4 & 50.4 & 40.6 \\
\quad + emb $k{=}20$ & 66.3 & 60.3 & 51.9 \\
\quad $\Delta$ & +8.9 & +9.9 & +11.3 \\
\midrule
Sonnet 4.5 flat & 66.8 & 57.0 & 45.9 \\
\quad + emb $k{=}20$ & 69.5 & 64.4 & 55.9 \\
\quad $\Delta$ & +2.7 & +7.4 & +10.0 \\
\bottomrule
\end{tabular}
\caption{Cross-model shortlisting recovery (F1 on implicit queries). GPT-5.4 at 325 tools is fold-averaged ($k{=}3$, $\sigma{=}2.0$); all other entries are fold~0. Sonnet on fold~0 only (API cost). Paired bootstrap 95\% CIs on full-scale deltas: GPT-5.4 [9.2, 11.5], GPT-5.1 [10.1, 12.5], Sonnet [8.9, 11.1]; all exclude zero.}
\label{tab:cross-model}
\end{table}

\subsection{Production Validation}
\label{sec:production-validation}

Table~\ref{tab:production} validates the synthetic findings on 1,435 human-annotated production utterances.
Despite flat baselines ranging from 28\% to 36\% F1 at full scale, all three models land within a 2pp band (44--46\%) once shortlisting is applied.

Absolute F1 is 10--15pp lower than synthetic, consistent with production traffic being entirely implicit and containing misspellings, fragments, and ambiguous intent absent from synthetic generation.
The synthetic-production gap is comparable for GPT-5.4 (14pp) and Sonnet (14pp), suggesting it reflects query difficulty rather than distributional affinity between GPT-4o query generation and GPT routing models.
At the query level, with shortlisting at full scale 80\% of synthetic queries and 60\% of production queries receive at least one correct tool.

\begin{table}[t]
\centering
\small
\begin{tabular}{lccc}
\toprule
& \textbf{120} & \textbf{325} & \textbf{584} \\
\textbf{Model} & \textbf{tools} & \textbf{tools} & \textbf{tools} \\
\midrule
GPT-5.4 flat & 42.2 & 36.8 & 27.9 \\
\quad + emb $k{=}20$ & 52.9 & 49.4 & 44.4 \\
\quad $\Delta$ & +10.7 & +12.7 & +16.5 \\
\midrule
GPT-5.1 flat & 50.3 & 45.0 & 36.2 \\
\quad + emb $k{=}20$ & 52.9 & 50.9 & 46.0 \\
\quad $\Delta$ & +2.6 & +6.0 & +9.8 \\
\midrule
Sonnet 4.5 flat & 53.6 & 41.9 & 32.2 \\
\quad + emb $k{=}20$ & 54.9 & 51.3 & 45.5 \\
\quad $\Delta$ & +1.3 & +9.4 & +13.3 \\
\bottomrule
\end{tabular}
\caption{Production validation: F1 on 1,435 human-annotated production utterances (implicit only). Scale 325 is fold-averaged ($k{=}3$); endpoints are single-fold. Shortlisting recovery replicates the synthetic pattern (Table~\ref{tab:cross-model}): +10--17pp at full scale, growing with catalog size. Paired bootstrap 95\% CIs on full-scale deltas exclude zero: [12.8, 20.2], [5.7, 14.0], [9.9, 16.8].}
\label{tab:production}
\end{table}

\subsection{Retrieval Method Comparison}
\label{sec:retrieval-comparison}

\paragraph{Tool-level beats pack-level.}
For both embedding models, tool-level retrieval ($k{=}20$) outperforms pack-level ($k{=}5$ packs/agents, expanded to $\sim$30 tools) by 2--4pp consistently, since pack expansion loads irrelevant sibling tools that dilute the candidate set. Part of this edge (${\sim}$2pp) comes from ranked retrieval exploiting the router's positional bias (Appendix~\ref{app:positional-bias}).
All pack-level approaches converge at full scale, with platform tool search (50.3\%), pack-level embedding (49.1\%), and hierarchical LLM routing (47.9\%) falling within 2.4pp at 584 tools.
The hierarchical baseline selects only 1.2 packs on average (83\% hit rate), making the pack intermediate step a source of unrecoverable error.

\paragraph{Large dense retriever outperforms lexical and fine-tuned alternatives.}
Text-embedding-3-large (52.5\%) outperforms ToolRet-e5 \cite{toolret}, a 335M retriever fine-tuned on 200k tool-retrieval pairs (48.7\%), by 4pp. The fine-tuned model was trained on API-centric data and may face an out-of-domain penalty on enterprise productivity tools. A domain-matched fine-tune at comparable scale could close or reverse this gap.
The fine-tuned retriever runs locally (${\sim}$2ms/query) vs.\ an API call (${\sim}$50ms).
BM25 shortlisting ($k{=}20$) falls below flat routing at every scale point (32.8\% vs.\ 42.1\% at 584 tools), as enterprise productivity tools share vocabulary across agents, making lexical matching ineffective.

\begin{figure*}[t]
\centering
\includegraphics[width=\textwidth]{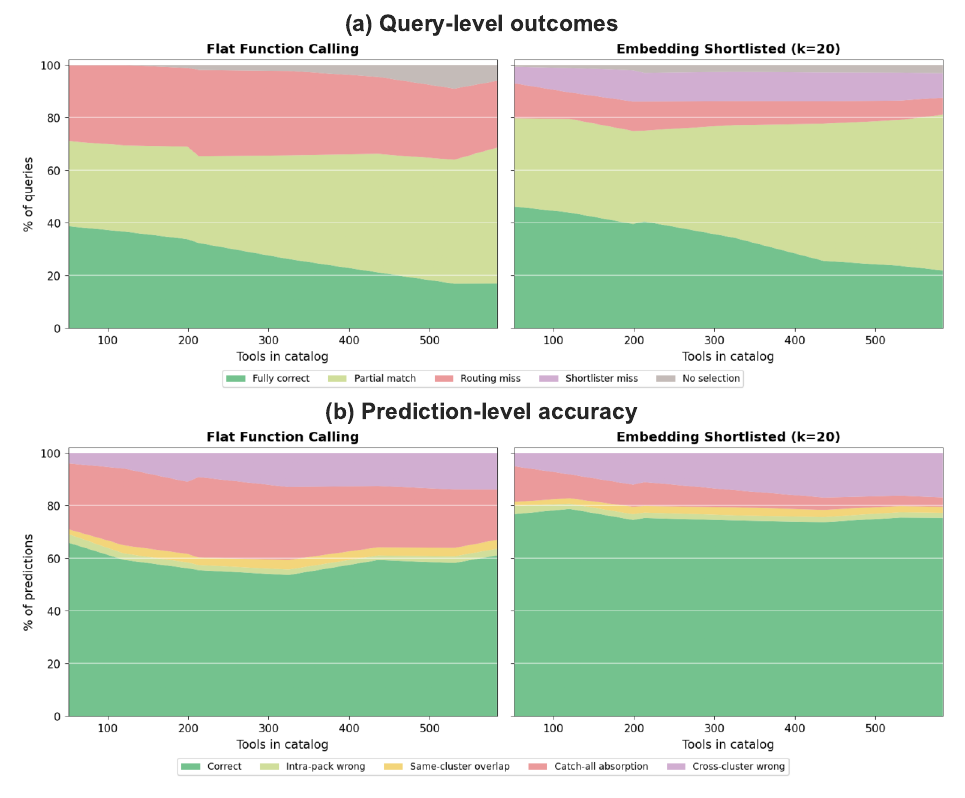}
\caption{Error analysis across catalog scale (GPT-5.4). (a)~Query-level outcomes: shortlisting converts routing misses (red) into correct predictions (green) at the cost of bounded shortlister misses (purple, ${\sim}$9\%). At full scale, 80\% of queries receive at least one correct tool with shortlisting vs.\ 69\% flat. (b)~Prediction-level accuracy: catch-all absorption (red) drops from 19\% to 4\% with shortlisting while cross-cluster confusion is largely unchanged.}
\label{fig:error-analysis}
\end{figure*}

\subsection{Error Analysis}
\label{sec:error-analysis}

\paragraph{Query outcomes.}
Figure~\ref{fig:error-analysis}a decomposes query outcomes into five categories.
As the catalog scales, fully correct predictions drop from 39\% to 17\% in flat routing and from 46\% to 22\% with shortlisting, while partial matches grow to fill the gap.
Shortlisting cuts routing misses from 31\% to 10\% at full scale at the cost of 9\% shortlister misses.
The trade is favorable: correct-or-partial coverage rises from 69\% to 80\%, and the shortlister miss rate is stable across scale.

\paragraph{Prediction-level accuracy.}
At the prediction level (Figure~\ref{fig:error-analysis}b), shortlisting improves accuracy from 61\% to 75\% correct at full scale.
The largest error in flat routing is catch-all absorption (general-purpose agents absorbing specific queries, 19\% of predictions), which shortlisting reduces to 4\%.
Cross-cluster confusion (routing to a wrong semantic cluster, 14\%$\rightarrow$17\%) and same-cluster overlap (routing to a similar tool in the correct cluster, ${\sim}$3\%) are largely unaffected, and intra-pack errors are negligible ($<$3\%).
Shortlisting helps by narrowing the candidate set, reducing the opportunity for general-purpose agents to absorb specific queries.

\paragraph{Positional bias in routing.}
The router exhibits primacy bias: ranked shortlisters provide $\sim$2pp of free accuracy from retrieval ordering, and oracle upper bounds with fixed ground-truth position overstate the routing ceiling by $\sim$4pp (Appendix~\ref{app:positional-bias}).

\section{Discussion}
\label{sec:discussion}

The retrieval gap (16pp) is addressable with better retrievers. Embedding shortlisting at $k{=}20$ adds ${\sim}$50ms (API) or ${\sim}$2ms (local) while shrinking the routing prompt from 584 to 20 tool definitions.
For catalogs beyond ${\sim}$30 agents (${\sim}$180 tools), the accuracy gain outweighs this cost.

The confusion gap (at least 10pp) is not recoverable by retrieval alone. Shortlisting narrows the candidate set but the router still confuses semantically similar tools.
Promising directions include description deduplication \cite{toolscope}, dynamic tool exclusion at routing time, and multi-turn clarification for ambiguous intent.

These findings are measured on a single enterprise productivity catalog with dense semantic overlap.
Domains with more functionally distinct tools may degrade slower, and the ${\sim}$30-agent threshold is catalog-specific.
The qualitative pattern (recall-driven degradation, recoverable by retrieval) is more likely to transfer than the specific numbers.

\section{Conclusion}
\label{sec:conclusion}

On a production catalog of 110 agents (584 tools), single-step routing degrades 16--23pp as the catalog scales.
The degradation is recall-driven and decomposes into a retrieval gap recoverable by better candidate selection and a confusion gap (at least 10pp on this catalog) that better retrieval alone cannot close.
Embedding shortlisting recovers +10--11pp F1 across three models and two providers, with tool-level retrieval consistently outperforming all pack-level approaches including platform tool search and hierarchical LLM routing.
A production annotation study (1,435 human-labeled utterances, three-way annotation) validates the synthetic findings, with shortlisting recovery replicating at +10--17pp on real traffic despite 10--15pp lower absolute performance.

\section*{Limitations}
\label{sec:limitations}

Our primary evaluation uses synthetic queries; the production validation (\S\ref{sec:production-validation}) confirms the pattern with fold-averaging at intermediate scale and moderate inter-annotator agreement ($\alpha{=}0.68$).
Production queries are entirely implicit and 10--15pp harder than synthetic, likely reflecting both distributional differences and annotation strictness.

All experiments use a single enterprise productivity catalog.
Domains with less semantic overlap (e.g., distinct API services) may degrade slower, while domains with more overlap (e.g., multiple coding assistants) may degrade faster.

GPT models use OpenAI function calling (Responses API); Sonnet uses Anthropic native tool use.
The cross-model comparison reflects both model and interface differences.
We verified that Sonnet's native tool use and text-in-prompt routing produce similar F1 ($<$1pp difference at scale 20), suggesting the interface effect is small for this model.

\bibliography{references}

\appendix

\section{Full Scaling Results by Model}
\label{app:full-results}

\begin{figure*}[t]
\centering
\includegraphics[width=\textwidth]{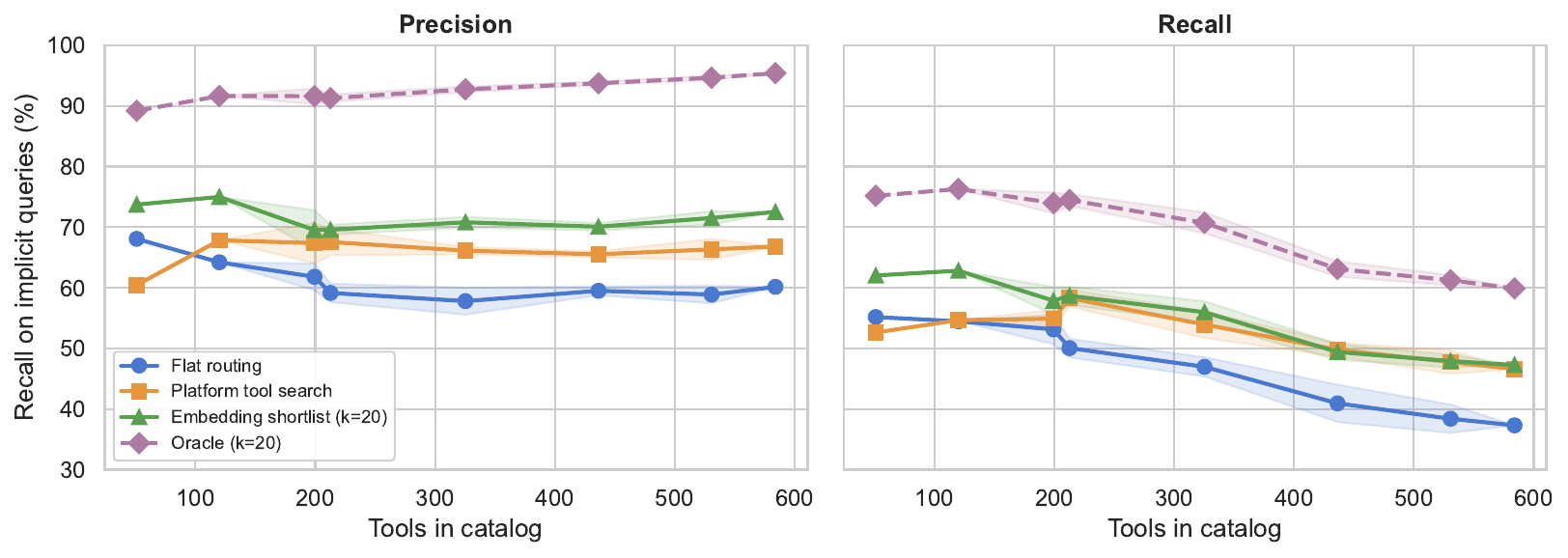}
\caption{Precision (left) and recall (right) on implicit queries across scale. Precision is relatively stable across scale. Recall drives the degradation, dropping 15pp even for the oracle.}
\label{fig:precision-recall}
\end{figure*}

Figure~\ref{fig:precision-recall} shows precision and recall curves across scale for all four approaches.
Tables~\ref{tab:full-gpt54}--\ref{tab:full-sonnet} report precision, recall, and F1 on implicit queries for flat routing and embedding shortlisting ($k{=}20$ tools, text-embedding-3-large) at all scale points.

GPT-5.4 and GPT-5.1 flat results are fold-averaged at intermediate scales ($k{=}3$ folds); all other entries are fold~0.
Scale~110 (584 tools) is the full catalog and has a single fold.

\begin{table}[h]
\centering
\small
\begin{tabular}{r rrr rrr}
\toprule
& \multicolumn{3}{c}{\textbf{Flat}} & \multicolumn{3}{c}{\textbf{Emb $k{=}20$}} \\
\cmidrule(lr){2-4} \cmidrule(lr){5-7}
\textbf{Tools} & \textbf{P} & \textbf{R} & \textbf{F1} & \textbf{P} & \textbf{R} & \textbf{F1} \\
\midrule
51   & 68.1 & 55.2 & 58.2 & 73.7 & 62.0 & 64.7 \\
120  & 64.2 & 54.5 & 56.1 & 75.0 & 62.8 & 65.4 \\
180  & 61.8 & 53.1 & 54.1 & 69.5 & 57.9 & 60.1 \\
230  & 59.1 & 50.0 & 51.3 & 69.6 & 58.7 & 60.6 \\
325  & 57.8 & 47.0 & 48.7 & 70.8 & 56.0 & 58.9 \\
430  & 59.5 & 40.9 & 44.3 & 70.1 & 49.4 & 53.2 \\
530  & 58.9 & 38.4 & 42.4 & 71.5 & 47.9 & 52.5 \\
584  & 60.2 & 37.3 & 42.1 & 72.5 & 47.3 & 52.5 \\
\bottomrule
\end{tabular}
\caption{GPT-5.4 scaling results (implicit queries). Intermediate scales are fold-averaged ($k{=}3$).}
\label{tab:full-gpt54}
\end{table}

\begin{table}[h]
\centering
\small
\begin{tabular}{r rrr rrr}
\toprule
& \multicolumn{3}{c}{\textbf{Flat}} & \multicolumn{3}{c}{\textbf{Emb $k{=}20$}} \\
\cmidrule(lr){2-4} \cmidrule(lr){5-7}
\textbf{Tools} & \textbf{P} & \textbf{R} & \textbf{F1} & \textbf{P} & \textbf{R} & \textbf{F1} \\
\midrule
51   & 76.5 & 58.8 & 63.7 & 80.8 & 63.8 & 68.6 \\
120  & 69.6 & 52.8 & 57.4 & 79.2 & 61.6 & 66.3 \\
180  & 69.1 & 53.2 & 57.3 & 76.7 & 59.7 & 64.2 \\
230  & 66.5 & 51.2 & 55.1 & 74.9 & 57.9 & 62.3 \\
325  & 64.3 & 45.7 & 50.4 & 76.9 & 55.5 & 60.3 \\
430  & 67.0 & 39.6 & 45.8 & 73.9 & 47.6 & 53.1 \\
530  & 65.7 & 36.8 & 43.0 & 74.8 & 44.9 & 51.1 \\
584  & 62.5 & 34.7 & 40.6 & 77.4 & 45.1 & 51.9 \\
\bottomrule
\end{tabular}
\caption{GPT-5.1 scaling results (implicit queries). Flat results are fold-averaged at intermediate scales. Embedding shortlisting on fold~0.}
\label{tab:full-gpt51}
\end{table}

\begin{table}[h]
\centering
\small
\begin{tabular}{r rrr rrr}
\toprule
& \multicolumn{3}{c}{\textbf{Flat}} & \multicolumn{3}{c}{\textbf{Emb $k{=}20$}} \\
\cmidrule(lr){2-4} \cmidrule(lr){5-7}
\textbf{Tools} & \textbf{P} & \textbf{R} & \textbf{F1} & \textbf{P} & \textbf{R} & \textbf{F1} \\
\midrule
51   & 79.1 & 61.8 & 66.3 & 80.4 & 64.2 & 68.6 \\
120  & 79.8 & 62.0 & 66.8 & 81.1 & 65.2 & 69.5 \\
180  & 76.7 & 60.8 & 64.6 & 79.7 & 65.8 & 69.3 \\
230  & 76.0 & 60.6 & 64.4 & 77.9 & 62.1 & 66.0 \\
325  & 70.8 & 51.3 & 56.2 & 79.0 & 60.3 & 64.4 \\
430  & 71.1 & 45.4 & 51.1 & 74.8 & 52.3 & 56.9 \\
530  & 69.8 & 41.9 & 48.1 & 75.8 & 50.2 & 55.3 \\
584  & 68.6 & 39.4 & 45.9 & 78.3 & 49.8 & 55.9 \\
\bottomrule
\end{tabular}
\caption{Claude Sonnet 4.5 scaling results (implicit queries). Flat intermediate scales are fold-averaged ($k{=}3$); all other entries are fold~0.}
\label{tab:full-sonnet}
\end{table}

\section{Retriever Comparison}
\label{app:retriever-comparison}

\begin{figure*}[t]
\centering
\includegraphics[width=\textwidth]{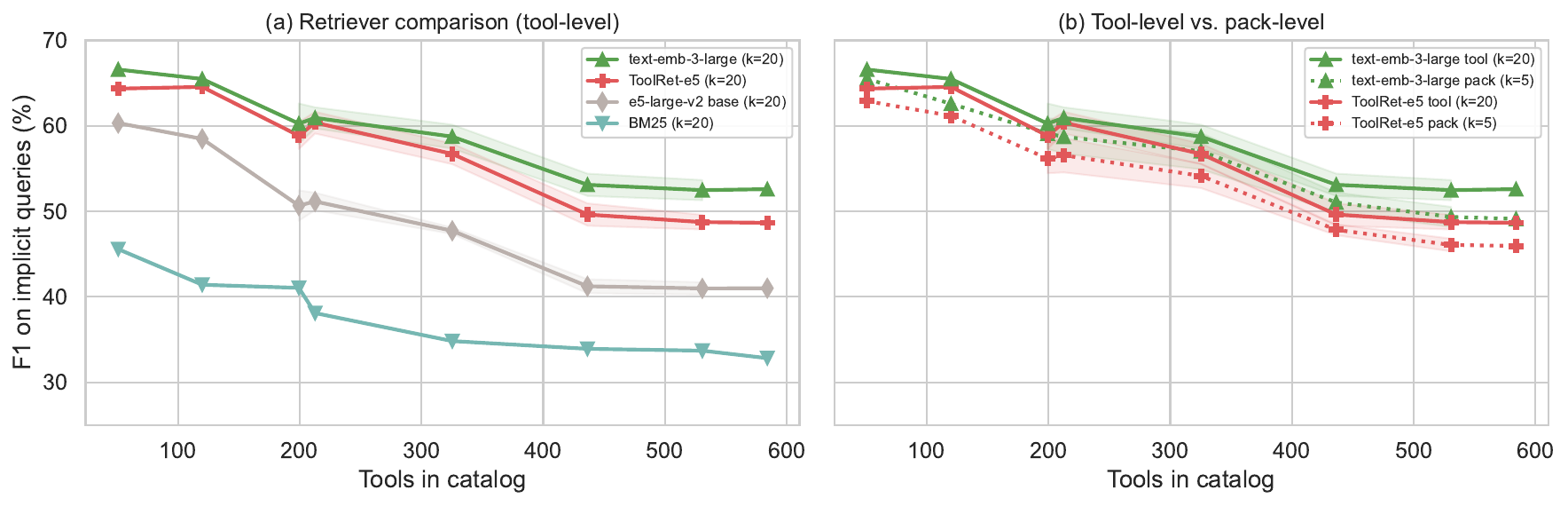}
\caption{Retriever comparison across scale (GPT-5.4). (a)~Tool-level retrievers ranked by end-to-end F1. General-purpose text-embedding-3-large outperforms fine-tuned ToolRet-e5 and base e5-large-v2. BM25 falls below flat routing (Figure~\ref{fig:unified-scaling}) at all scales. (b)~Same retriever, tool-level (solid) vs.\ pack-level (dotted). Tool-level consistently wins by 2--4pp.}
\label{fig:retriever-comparison}
\end{figure*}

Figure~\ref{fig:retriever-comparison} isolates retrieval method and granularity effects.
Panel~(a) compares four tool-level retrievers. BM25 and base e5-large-v2 both fall below flat routing, demonstrating that low-quality retrieval is counterproductive.
Panel~(b) shows that tool-level retrieval outperforms pack-level by 2--4pp for both embedding models.

\section{K-Sensitivity}
\label{app:k-sensitivity}

F1 plateaus at $k{\geq}10$ tools and is statistically indistinguishable from $k{=}20$ to $k{=}50$ (bootstrap $p{=}0.78$ for $k{=}20$ vs.\ $k{=}35$).
We fix $k{=}20$ for all scaling experiments as the cheapest point on the plateau (93\% embedding recall at 3.4\% of catalog).

\section{Fixed-Cohort Validation}
\label{app:fixed-cohort}

To confirm degradation is not driven by harder queries entering the pool at larger scales, we track a fixed cohort of 731 implicit queries whose target tools are present at every scale point (fold~0).
On this cohort, flat-routing F1 drops from 58.2\% at 51 tools to 43.3\% at 584 tools, a 14.9pp degradation on \emph{identical queries} as the catalog grows.
Embedding shortlisting partially recovers this (46.4\% at 584 tools, $+$3.1pp), confirming both the degradation and the recovery are genuine.

\section{Pack-Level Approaches}
\label{app:tool-search}
\label{app:hierarchical}

\paragraph{Platform tool search.}
OpenAI's namespace-based tool search exhibits a crossover effect: $-$6.3pp at 10 agents, $+$7--9pp from 30 agents onward.
At small scale, namespace selection adds unnecessary indirection. At larger scale, namespace filtering reduces the option space.
The ceiling near 60 agents reflects within-namespace tool \emph{selection} degrading 14pp, nearly double the namespace retrieval degradation.

\paragraph{Hierarchical LLM routing.}
Two-stage routing (LLM selects pack, then routes within pack) achieves 47.9\% F1 at full scale, below all other shortlisting approaches.
The LLM selects only 1.2 packs on average (83\% hit rate) despite a recall-oriented prompt, making the pack decision a source of unrecoverable error.

\section{Positional Bias in Routing}
\label{app:positional-bias}

Embedding shortlisting returns candidates ranked by similarity, placing the correct tool near the top.
To isolate the filtering benefit from any ranking advantage, we re-run all scale points with shuffled candidate order (per-query deterministic permutation).

\begin{table}[h]
\centering
\small
\begin{tabular}{lcc}
\toprule
\textbf{Condition} & \textbf{Avg $\Delta$ vs flat} & \textbf{Position effect} \\
\midrule
Ranked (default) & +8.8pp & --- \\
Shuffled & +6.8pp & 2.0pp \\
\midrule
Oracle ranked & +25.8pp & --- \\
Oracle shuffled & +22.0pp & 3.8pp \\
\bottomrule
\end{tabular}
\caption{Positional bias decomposition (implicit F1, averaged across all scale points). Embedding ranking contributes $\sim$2pp; oracle ranking contributes $\sim$4pp because ground truth is always at rank~1. All deltas are statistically significant (paired bootstrap, $p < 0.01$).}
\label{tab:positional-bias}
\end{table}

The router favors candidates presented earlier in the tool list.
For embedding shortlisting, this accounts for ${\sim}$2pp of the total gain, stable across scale points.
For oracle shortlisting, the effect is larger (${\sim}$4pp) because ground truth is always at rank 1, meaning oracle upper bounds overstate the routing ceiling.

\end{document}